\documentclass[letterpaper]{article} 
\usepackage[submission]{aaai24}  
\usepackage{times}  
\usepackage{helvet}  
\usepackage{courier}  
\usepackage[hyphens]{url}  
\usepackage{graphicx} 
\usepackage{natbib}  
\usepackage{caption} 
\usepackage{algorithm}
\usepackage{algorithmic}

\usepackage{booktabs}
\usepackage{multirow}
\usepackage{subcaption}
\usepackage{amsmath}
\usepackage{cleveref}

\usepackage{newfloat}
\usepackage{listings}
\DeclareCaptionStyle{ruled}{labelfont=normalfont,labelsep=colon,strut=off} 
\floatstyle{ruled}
\newfloat{listing}{tb}{lst}{}
\floatname{listing}{Listing}
\title{Defending Label Inference Attacks in Split Learning under Regression Setting}

\author{
Haoze Qiu$^1$ \and
Fei Zheng$^1$ \and
Chaochao Chen$^1$ \and
Xiaolin Zheng$^1$
\affiliations
$^1$Zhejiang University \\
\emails
\{22221304, zfscgy2, zjuccc, xlzheng\}@zju.edu.cn}

\begin{document}

\maketitle

\begin{abstract}
As a privacy-preserving method for implementing Vertical Federated Learning, Split Learning has been extensively researched.
However, numerous studies have indicated that the privacy-preserving capability of Split Learning is insufficient.
In this paper, we primarily focus on label inference attacks in Split Learning under regression setting, which are mainly implemented through the gradient inversion method.
To defend against label inference attacks, we propose Random Label Extension (RLE), where labels are extended to obfuscate the label information contained in the gradients, thereby preventing the attacker from utilizing gradients to train an attack model that can infer the original labels.
To further minimize the impact on the original task, we propose Model-based adaptive Label Extension (MLE), where original labels are preserved in the extended labels and dominate the training process.
The experimental results show that compared to the basic defense methods, our proposed defense methods can significantly reduce the attack model's performance while preserving the original task's performance.
\end{abstract}

\section{Introduction}
With the growing concern for privacy protection in machine learning, Split Learning (SL) ~\cite{vepakomma2018split,li2021label} has been widely researched and applied as a privacy-preserving technology for implementing Vertical Federated Learning (VFL).
Compared to joint encrypted computation implemented through cryptographic methods ~\cite{mohassel2017secureml,huang2022cheetah}, Split Learning involves much lower computational costs and higher efficiency.
In the typical two-party Split Learning scenario, the Split Learning system is divided into a feature party and a label party.
The label party exclusively holds all the labels, while the feature party holds all the feature data except for the labels.
Additionally, the label party controls the top model, while the feature party controls the bottom model.
%

The process of Split Learning can be divided into forward propagation and backward propagation (For convenience, we describe the commonly used notations in \Cref{table:notations}): 
\begin{itemize}
    \item Forward propagation: the feature party feeds the features $X$ to $M_b$, resulting in the output of the bottom model $E = M_b(X, W_b)$. Subsequently, the feature party transmits $E$ to the label party, who then feeds $E$ into $M_t$ to obtain the final prediction result $\hat Y = M_t(E, W_t)$.
    \item Backward propagation: After obtaining $\hat Y$, the label party calculates the loss $L(\hat Y, Y)$, the gradients $G_E = \partial L(\hat Y, Y)/\partial E$ and $G_t = \partial L(\hat Y, Y)/\partial W_t$. $G_t$ is directly utilized to update $W_t$, while $G_E$ is transmitted to the feature party. According to the chain rule, by multiplying $G_E$ with the gradient $\partial E/\partial W_b$, the feature party obtains the gradient $G_b = \partial L(\hat Y, Y)/\partial W_b$. Finally, the gradient $G_b$ is utilized to update $W_b$.
\end{itemize}

Split Learning is efficient, but its privacy protection ability is limited.
Numerous studies have proposed feature inference attacks ~\cite{abuadbba2020can,luo2021feature} or label inference attacks ~\cite{aggarwal2021label,fu2022label,liu2022clustering} for the classification problem in Split Learning.
Instead, we focus on label inference attacks in Split Learning under regression setting ~\cite{xie2023label} in this paper.
During the training process of Split Learning, there exists a certain binding relationship between the ground-truth labels and the gradients transmitted from the label party to the feature party.
Therefore, the feature party, as the attacker, can utilize the trained bottom model and the received gradients to train an attack model that can infer ground-truth labels.
Additionally, due to the impracticality of obtaining a large amount of leaked labels, the model completion method (fine-tune the attack model with leaked labels) is used as an auxiliary method to improve the overall performance of the attack model.
%

The existence of such privacy vulnerabilities in Split Learning is unacceptable, thus it is urgent to investigate corresponding defense strategies.
Applying cryptographic methods is a solution to privacy leakage issues in Split Learning.
However, cryptographic methods involve high computational costs, making them impractical for real-world applications.
Another solution is to perturb the shared information or the sensitive data. 
~\cite{fu2022label,xie2023label} have explored defense methods such as gradient compression and adding noise to gradients or sensitive data.
However, these methods cannot balance defense effect and preservation of the original task's performance.
%

In order to overcome the limitations of existing defense methods, we propose a novel defense method: Label Extension.
We find that extending original labels can obfuscate the label information contained in the gradients, thus preventing the attacker from using the gradient inversion method to train the attack model.
Firstly, we propose the Random Label Extension (RLE) method, which simply extends the original labels with random vectors.
Besides, to maintain the original task's performance, we preserve original label information in the extended $D$-dimensional labels.
Experimental results indicate that the RLE method can effectively defend against label inference attacks, and its ability to preserve the original task's performance is superior to that of basic perturbation-based defense methods.
Considering that the RLE method still has the potential to impact the original task’s performance to some extent, we further propose the Model-based adaptive Label Extension (MLE) method.
In MLE, the dimension where the original labels
are located is designed to dominate the training process, with the aim of enhancing the original task’s performance when defense method is applied.
Experimental results indicate that the MLE method can effectively defend against label inference attacks with a negligible impact on the original task’s performance.
Totally, we make the following main contributions:
\begin{itemize}
    \item We investigate various defense methods against label inference attacks in Split Learning under regression setting, including adding noise to the labels, adding noise to the gradients, and gradient compression.
    \item We propose the Random Label Extension (RLE) method and the Model-based adaptive Label Extension (MLE) method to solve the problem of degrading the original task's performance when applying basic perturbation-based defense methods.
    \item Experiments on multiple datasets demonstrate that our proposed defense methods can significantly reduce the attack model’s performance with little to no impact on the original task’s performance.
\end{itemize}
\begin{table}[t]
\footnotesize
    \centering
    \begin{tabular}{cl}
    \toprule
        Notation       & Definition    \\
    \midrule
        $M_b, M_t, M_s$    & Bottom Model, Top Model, Surrogate Model \\
        $W_b, W_t, W_s$    & The weight of $M_b, M_t, M_s$ \\
        $X$                & Input instance \\
        $E$                & Output of $M_b$ \\
        $\hat Y, \hat Y^*$ & Model prediction, dummy model prediction \\
        $Y, Y^*$           & Ground-truth labels, dummy labels \\
        $Y_{LE}$           & Extended labels \\
        $L(\hat Y, Y)$     & Original loss function \\
        $L(\hat Y^*, Y^*)$ & Dummy loss function \\
        $G_E$              & Gradient ($\partial L(\hat Y, Y)/\partial E$) \\
        $G_E^*$            & Dummy gradient ($\partial L(\hat Y^*, Y^*)/\partial E$) \\
        $D_E$              & Dimension of $E$ \\
        $D$                & Dimension of the extended labels \\
        $n$                & Number of samples \\
    \bottomrule
    \end{tabular}
    \caption{Notations \& Definitions.}
    \label{table:notations}
\end{table}

\section{Related Work}
\subsection{Privacy Attacks Against Split Learning}
Many studies have investigated privacy attacks against Split Learning.
~\cite{abuadbba2020can} first introduce the concept of feature inference attacks in Split Learning. They find that when Split Learning is applied on 1d CNN models, the attacker can infer input features based on the intermediate output.
~\cite{luo2021feature,pasquini2021unleashing} propose feature inference attacks for different Split Learning scenarios and verify their effectiveness through experiments.
~\cite{li2021label} first introduce the concept of label inference attacks in Split Learning. They utilize the difference of gradients to implement label inference attacks in binary classification problems.
\cite{sun2022label,liu2022clustering} propose utilizing forward embedding to infer labels in multi-class classification problems.
\cite{fu2022label} propose using leaked labels to fine-tune the attack model to infer labels, while \cite{zou2022defending} propose using gradients to train the attack model to infer labels, both for multi-classification problems in Split Learning.
~\cite{xie2023label} propose implementing label inference attacks in Split Learning for regression problems by combining the gradient inversion method and the model completion method.

\subsection{Privacy Protection for Split Learning}
One approach to protect privacy in Split Learning is to use carefully designed cryptographic protocols.
~\cite{mohassel2017secureml,rathee2020cryptflow2,fu2022blindfl,huang2022cheetah} employ various cryptographic modules protect the privacy of the training process.
However, the computational costs of cryptographic methods are too heavy to be practical in many Split Learning scenarios.
Another approach to protect privacy is to perturb the shared information or the sensitive data.
~\cite{vepakomma2020nopeek} propose the distance correlation loss to defend against feature inference attacks by perturbing the correlation between the input features and the output of the bottom model.
~\cite{zheng2022making} propose the potential energy loss to defend against label inference attacks in classification problems by pushing outputs of the same class toward the decision boundary.
~\cite{ghazi2021deep,wu2022does,xie2023label} attempt to protect privacy information by adding noise to raw data or gradients.
~\cite{zou2022defending} attempt to reduce the privacy information contained in gradients by applying gradient compression.

\section{Defense Against Label Inference Attacks}
We explain the specific implementation for label inference attacks in section 3.1 and describe the basic defense methods and our proposed defense methods in section 3.2.
This paper primarily focuses on the practical Split Learning scenario with two participating parties, as illustrated in \Cref{fig:attack architecture}, where the labels in the regression tasks are single-dimensional.
Actually, the attack and defense methods presented in this paper can be naturally extended to scenarios involving multiple participating parties, where the labels are multi-dimensional.

\subsection{Label Inference Attacks}
\label{sec:attack}
The primary objective of this section is to present the implementation method of label inference attacks.
The feature party without access to ground-truth labels is regarded as the attacker, while the label party is considered the victim.
According to~\cite{xie2023label}, label inference attacks in Split Learning under regression setting are mainly implemented through the gradient inversion method, with the assistance of the model completion method.
\begin{figure}[t]
    \centering
    \includegraphics[width=1\linewidth]{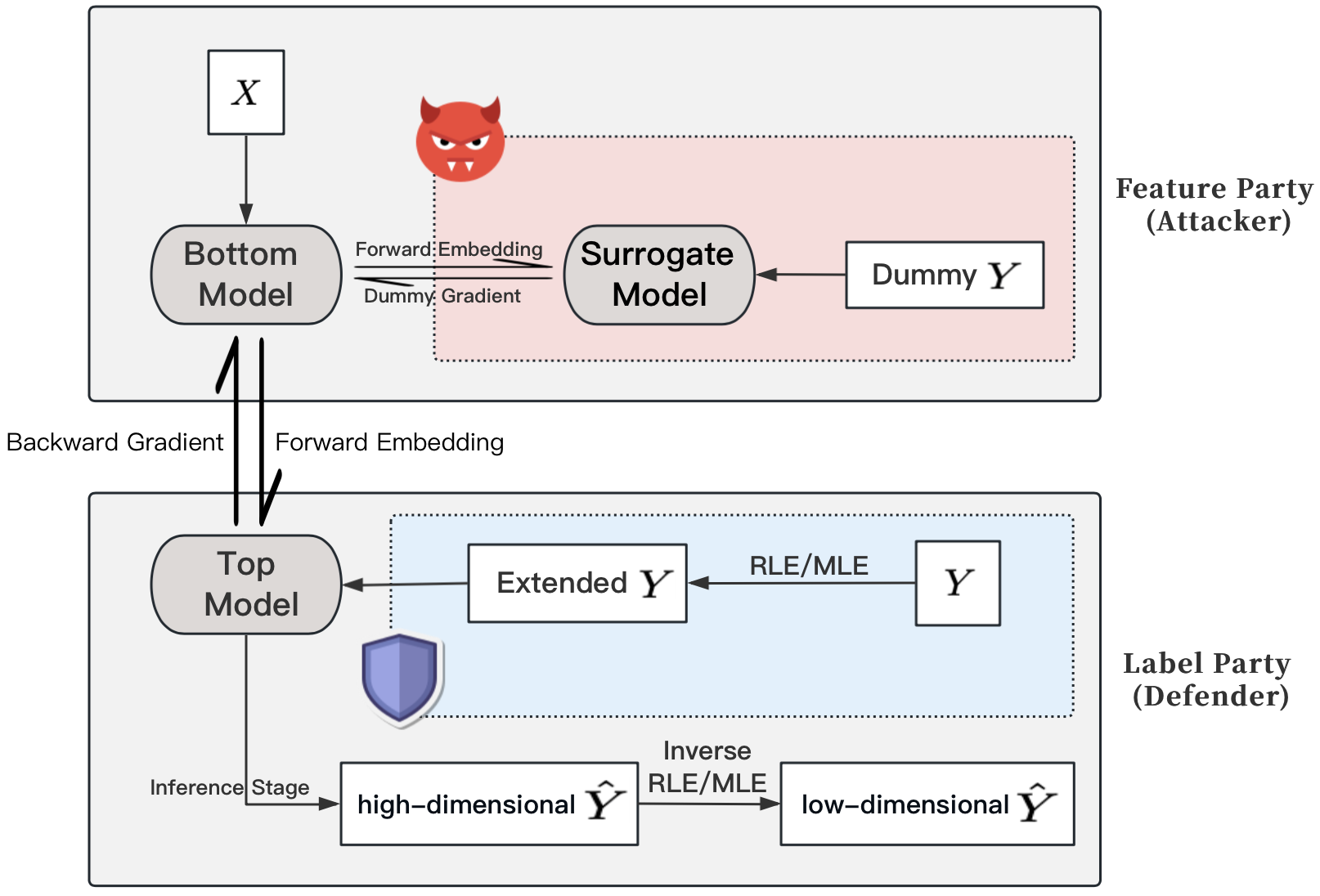}
    \caption{Overview of the attack and defense framework.The feature party intends to conduct label inference attacks, while the label party defends against label inference attacks.}
    \label{fig:attack architecture}
\end{figure}
Specifically, the feature party first constructs a surrogate model $M_s$ to replace the top model $M_t$ and constructs dummy labels $Y^*$ to replace the inaccessible ground-truth labels $Y$. Meanwhile, the feature party keeps the trained bottom model $M_b$ fixed throughout the process. 
Subsequently, the feature party utilizes the gradients received from the label party during training, as well as leaked labels, to train and update the randomly initialized $M_s$ and $Y^*$.
When $M_s$ and $Y^*$ are trained to their optimal state, dummy labels $Y^*$ can be considered to be in close proximity to the original labels $Y$.
The complete model obtained by combining the updated $M_s$ and the fixed $M_b$ is capable of predicting the labels of arbitrary samples.
The attack framework is illustrated in \Cref{fig:attack architecture}.
The subsequent paragraphs will provide explanations of the gradient inversion method and the model completion method employed in label inference attacks.
\textbf{Method: Gradient Inversion.}
~\cite{zhu2019deep} first propose utilizing the gradient inversion method to steal private information in Federated Learning.
~\cite{zou2022defending} apply the gradient inversion method to achieve label inference attacks for the classification tasks in Vertical Federated Learning.
In Split Learning under regression setting, the label inference attacks are also mainly achieved through the gradient inversion method.
During the training process, the feature party saves the gradient $G_E$ received from the label party. 
Upon the completion of the training process, the feature party constructs $M_s$ and generates $Y^*$.
Subsequently, the feature party fixes the trained bottom model $M_b$ and simulates the process of Split Learning.
The feature party feeds the output of the fixed bottom model $E$ into $M_s$, which produces the dummy prediction $\hat Y^*$ as the output.
Afterwards, the dummy loss between the dummy model prediction and the dummy label $L(\hat Y^*, Y^*)$ is calculated, as well as the dummy gradient $G_E^* = \partial L(\hat Y^*, Y^*) / \partial E$.
In order to maximize the similarity between $Y^*$ and $Y$, and to ensure that the function of $M_s$ approximates the function of $M_t$, the dummy gradient $G_E^*$ needs to be optimized to closely align with the original saved gradient $G_E$.
The loss function for the gradient inversion method is as follows, where the second term is introduced to prevent multiple solutions.
\begin{small}
\begin{equation}
\begin{split}
    L_{gi} = L(\frac{\partial L(\hat Y, Y)}{\partial E}, \frac{\partial L(\hat Y^*, Y^*)}{\partial E}) + L(M_s(E), Y^*).\\
\end{split}
\end{equation}
\end{small}
\textbf{Method: Model Completion.}
~\cite{fu2022label} first propose label inference attacks implemented through model completion.
They suggest utilizing leaked labels to train the surrogate model $M_s$ for classification tasks in Split Learning.
The fine-tuned $M_s$ is then combined with the fixed bottom model $M_b$ to form a complete prediction model, which is capable of predicting labels of arbitrary samples.
However, under regression setting in Split Learning, label inference attacks implemented through model completion require a significant amount of leaked labels to achieve a promising attack performance.
In reality, it is not feasible to obtain too many real labels. Therefore, under regression setting in Split Learning, we use the model completion method as an auxiliary attack method to improve attack performance.
It is assumed that the feature party has access to a certain quantity of ground-truth labels $Y_{leaked}$ corresponding to a set of samples $X_{leaked}$.
The feature party fixes the trained bottom model $M_b$ and fine-tunes $M_s$ using $X_{leaked}$ and $Y_{leaked}$.
The loss function for the model completion method is as follows.
\begin{equation}
\begin{split}
    L_{mc} = L(M_s(M_b(X_{leaked})), Y_{leaked}). \\
\end{split}
\end{equation}
Besides, the parameter $\alpha$ is introduced to combine the two losses and obtain the overall loss for the attack algorithm.
\begin{equation}
\begin{split}
\label{equation:loss}
    L_{attack} = L_{gi} + \alpha \cdot  L_{mc}.\\
\end{split}
\end{equation}
After calculating the attack loss $L_{attack}$, the gradient of the attack loss $L_{attack}$ with respect to the weight of the surrogate model $W_s$ and dummy label $Y^*$ are calculated. 
$W_s$ and $Y^*$ are then updated for multiple iterations according to the learning rate $\eta$.
\begin{equation}
\begin{split}
    W_s = W_s -  \eta \cdot \frac{\partial L_{attack}}{\partial W_s},\\
    Y^* = Y^* -  \eta \cdot \frac{\partial L_{attack}}{\partial Y^*}.\\
\end{split}
\end{equation}

\subsection{Defense}
\label{sec:defense}
In Split Learning under regression setting, the objective of the feature party in executing label inference attacks is to obtain the ground-truth labels $Y$ corresponding to the samples $X$.
To mitigate label inference attacks, the label party may opt to perturb the original labels or the gradients.
It is possible to achieve the objective of reducing the attack performance by utilizing basic defense methods such as adding noise and gradient compression.
However, these methods inevitably compromise the original task's performance while effectively defending against label inference attacks.
In order to defend against label inference attacks without compromising the original task's performance, we propose a novel defense method: Random Label Extension.
Similar to adding noise to the original labels, the random label extension method also involves perturbing the original labels.
However, compared to directly adding noise to the original labels, the random label extension method preserves the original labels in the extended labels.
As illustrated in \Cref{fig:attack architecture}, the label party replaces the original labels with the extended labels during the training process, and obtains the $t$-th dimension of the $D$-dimensional model output as the final prediction during the inference process.
To further minimize the impact on the original task, we propose Model-based adaptive Label Extension, where the extended labels are derived from the $D$-dimensional outputs of the current model.
We will provide explanations of the basic defense methods and our proposed defense methods in this section.

\subsubsection{Basic Defense Methods}
We mainly investigate typical defense methods such as adding noise and gradient compression.
Both of these defense methods involve a tradeoff between the original task's performance and the effectiveness of the defense.
The following paragraphs provide an introduction to these two basic defense methods respectively.
\textbf{Adding Noise.}
~\cite{dwork2006calibrating,dwork2014algorithmic} introduce the concept of differential privacy, and further explore its practical applications in different scenarios.
~\cite{wei2020federated,hu2020personalized,chen2022differential} apply differential privacy to enhance the security of federated learning systems.
Therefore, it is natural to consider the widely used defense mechanism of adding noise to mitigate label inference attacks.
As the feature party intends to infer the labels held by the label party, the label party can protect the labels by injecting noise directly into them.
Given that the feature party infers the labels by gradients $G_E$, the label party can also protect the labels by adding noise to the gradients $G_E$ that are transmitted to the feature party.
We denote the noised label and gradient as $y_n$ and $g_n$, respectively, and denote noise distributions as $\mathcal{N}_y$ and $\mathcal{N}_g$, e.g., Gaussian or Laplace distributions.
\begin{equation}
\begin{split}
    y_n = y   + n_y \quad n_y\sim\mathcal{N}_y, \\
    g_n = g + n_g \quad n_g\sim\mathcal{N}_g. \\
\end{split}
\end{equation}
However, it remains challenging to strike a balance between the original task's performance and the defense effect, regardless of whether the noise is added to labels or gradients.
When adding large-scale noise, the original task's performance can be substantially reduced, which renders the defense meaningless; when adding small-scale noise, the defense effect may not be significant enough.
\textbf{Gradient Compression.}
~\cite{chen2021communication,castiglia2022compressed,zheng2023reducing} propose compressing the gradients $G_E$ using methods such as quantization and sparsification to improve communication efficiency in Split Learning.
Since the feature party infers the labels by gradients $G_E$, compressing the gradients can not only reduce the amount of transmitted data, but also provide effective defense against label inference attacks.
Therefore, we consider gradient compression as another basic defense method against label inference attacks.
Similarly, balancing between maintaining high performance for the original task and providing effective defense against attacks is a challenging task for gradient compression.

\subsubsection{RLE: Random Label Extension}
Considering the limitations of basic defense methods, it is imperative to explore advanced defense methods that can effectively defend against label inference attacks while maintaining high performance for the original task.
Since the attacker, i.e., the feature party, aims to infer the ground-truth labels, we still choose to perturb the labels.
Adding noise directly to the original labels can significantly affect the original task's performance while defending against the attacks. Therefore, we consider preserving the original label information in the perturbed labels.
We propose a novel method for label perturbation: Label Extension.
In our proposed method, the original single-dimensional labels are extended to $D$-dimensional vectors, where the original labels are located in the $t$-th dimension and the other dimensions can be considered as noise. 
The label extension method perturbs the original labels by extending them with noise, rather than adding noise directly to them, thereby preserving the original label information while adding noise.
According to \Cref{fig:attack architecture}, during the training process of the original task, the label party replaces the original single-dimensional labels with the extended $D$-dimensional labels to prevent the feature party from inferring the original labels using the received gradients.
And the model's output dimension is adjusted to $D$ as well.
During the inference process of the original task, the label party obtains the $t$-th dimension of the $D$-dimensional model outputs as the final prediction result, ensuring that the original task's performance is not significantly impacted despite the label perturbation.
At the outset, we propose a simple and straightforward defense method: Random Label Extension.
In the random label extension method, we simply extend the original labels with random vectors $y'_i = [r_1, r_2, \cdots, r_D], r_1, \cdots, r_D \sim \mathcal N(0, {\sigma}^2), i=1,2,...n$.
\begin{equation}
\begin{split}
    Y_{LE} = [y'_1, y'_2, \cdots, y'_n]. \\
\end{split}
\end{equation}
\begin{equation}
\begin{split}
    {Y_{LE}}_{:,t} = Y. \\
\end{split}
\end{equation}
$Y_{LE}$ denotes the extended labels, where the $t$-th dimension corresponds to the original labels.
Additionally, each dimension of the random vectors satisfies a certain distribution, e.g., the Gaussian distribution. 
In this paragraph, we will demonstrate the effectiveness of the random label extension method against label inference attacks.
Mean squared error is employed as the loss function in this context.
During the proof, we consider the scenario where the top model consists of a single linear layer. When the top model consists of multiple layers, a similar approach can be utilized to demonstrate the effectiveness.
We consider all scalar equations for a single sample, where $w_{ij}$ denotes an element in the weight matrix of the top model, $(b_1, b_2, ..., b_j, ..., b_D)$ denotes the bias of the top model, $(e_1, e_2, ..., e_i, ..., e_{D_E})$ denotes the output of the bottom model, $(\hat y^*_1, \hat y^*_2, ..., \hat y^*_j, ..., \hat y^*_D)$ denotes the dummy model prediction:
\begin{small}
\begin{equation}
\begin{split}
    & \hat y^*_1 = w_{11} \cdot e_1 + w_{21} \cdot e_2 + ... + w_{D_E1} \cdot e_{D_E} + b_1, \\
    & \hat y^*_2 = w_{12} \cdot e_1 + w_{22} \cdot e_2 + ... + w_{D_E2} \cdot e_{D_E} + b_2, \\
    & \cdots  \\
    & \hat y^*_D = w_{1D} \cdot e_1 + w_{2D} \cdot e_2 + ... + w_{D_ED} \cdot e_{D_E} + b_D. \\
\end{split}
\end{equation}
\end{small}
And $(g_1, g_2, ..., g_i, ..., g_{D_E})$ denotes the received gradient, $(y^*_1, y^*_2, ..., y^*_j, ..., y^*_D)$ denotes the dummy label: 
\begin{small}
\begin{equation}
\begin{split}
    & g_1 = \frac{\partial [(\hat y^*_1 - y^*_1)^2 + (\hat y^*_2 - y^*_2)^2 + ... + (\hat y^*_D - y^*_D)^2]}{\partial e_1}, \\
    & g_2 = \frac{\partial [(\hat y^*_1 - y^*_1)^2 + (\hat y^*_2 - y^*_2)^2 + ... + (\hat y^*_D - y^*_D)^2]}{\partial e_2}, \\
    & \cdots  \\
    & g_{D_E} = \frac{\partial [(\hat y^*_1 - y^*_1)^2 + (\hat y^*_2 - y^*_2)^2 + ... + (\hat y^*_D - y^*_D)^2]}{\partial e_{D_E}}. \\
\end{split}
\end{equation}
\end{small}
The feature party relies on these equations to infer the original labels. 
In the above equations, $e_i$, $g_i$ are known quantities, while $w_{ij}$, $b_j$, $y^*_j$ are unknown quantities.
Since the number of samples is $n$, the number of scalar unknowns is $D+D*D_E+D*n$, and the number of scalar equations is $D_E*n$.
When $D \geq D_E$, the number of scalar equations is less than the number of scalar unknowns, leading to the theoretical possibility of existing no unique solution.
Therefore, provided that the original labels are extended to at least $D_E$ dimensions in the random label extension method, the feature party is unable to complete label inference attacks.

\subsubsection{MLE: Model-based adaptive Label Extension}
\begin{algorithm}[tb]
    \caption{Split Learning with RLE/MLE protection}
    \label{alg:RLE/MLE}
    \textbf{Input}: samples $X$; untrained $M_b, M_t$; ground-truth labels $Y$ \\
    \textbf{Output}: trained $M_b, M_t$
    \begin{algorithmic}[1]
        \STATE the label party initializes the extended labels: \\
        $y'_i = [r_1, \cdots, r_j, \cdots, r_D], r_j \sim \mathcal N(0, {\sigma}^2)$ \\
        $i=1,2,\cdots,n \quad j=1,2,\cdots,D$\\
        $Y_{LE} = [y'_1, \cdots, y'_i, \cdots, y'_n], {Y_{LE}}_{:,t} = Y$;
        \FOR{$epoch = 1,2,3,...$} {
            \STATE the feature party calculates $E = M_b(X)$ \\
            and transmits $E$ to the label party;
            \IF{MLE is applied} {
                \STATE the label party calculates new extended labels: \\
                $Y_{LE} = M_t(E), {Y_{LE}}_{:,t} = Y$;
            }
            \ENDIF
            \STATE the label party calculates the model prediction and the loss: $\hat Y= M_t(E), L = MSE(\hat Y, Y_{LE})$;
            \STATE the label party calculates $\frac{\partial L}{\partial W_t}$ and updates $W_t$;
            \STATE the label party calculates $\frac{\partial L}{\partial E}$ and transmits it to the feature party;
            \STATE the feature party calculates $\frac{\partial L}{\partial W_b}$ and updates $W_b$;
        }
        \ENDFOR
    \end{algorithmic}
\end{algorithm}
The random label extension method is effective in resisting label inference attacks and can better preserve the original task's performance compared to basic defense methods.
However, due to the randomness of the extended vectors, the random label extension method still has the potential to reduce the original task's performance to some extent.
To further minimize the impact of the defense method on the original task, we propose the improved defense method: Model-based adaptive Label Extension.
In the model-based adaptive label extension method, the extended labels are derived from the $D$-dimensional outputs of the current model.
Specifically, in each iteration during the training process, the label party feeds the output of the bottom model $E$ into the current top model $M_t$. 
The extended labels for an iteration is obtained by replacing the $t$-th dimension of the $D$-dimensional output of the current top model with the original labels: 
\begin{equation}
\begin{split}
    Y_{LE} = M_t(E). \\
\end{split}
\end{equation}
\begin{equation}
\begin{split}
    {Y_{LE}}_{:,t} = Y. \\
\end{split}
\end{equation}
During each epoch of training, the loss is generated only by the $t$-th dimension where the original labels are located, while the loss generated by the other dimensions is zero.
Therefore, the $t$-th dimension where the original labels are located dominates the training process, which theoretically enhances the preservation of the original task's performance.
Additionally, in the model-based adaptive label extension method, the labels are still extended to at least $D_E$ dimensions, hence the same proof as the random label extension method can show that the defense against label inference attacks is effective.
The experimental results in the next chapter further demonstrate that the model-based adaptive label extension method can achieve a defense effect similar to that of the random label extension method while hardly affecting the original task's performance.
\Cref{alg:RLE/MLE} describes the Split Learning algorithm with RLE or MLE.

\section{Experiments}
\begin{table}[t]
    \centering
    \resizebox{\linewidth}{!}{
        \begin{tabular}{ccccc}
            \toprule
            Dataset & Size & Features & $M_b$ & $M_t$ \\
            \midrule
            Boston Housing & 506 & 13 & FC-2 & FC-1 \\
            California Housing & 20,640 & 8 & FC-3 & FC-2 \\
            Power Plant & 9,568 & 4 & FC-3 & FC-2 \\
            \bottomrule
        \end{tabular}
    }
    \caption{Datasets and Models.}
    \label{table:datasets}
\end{table}
\begin{table*}[t]
    \centering
    \resizebox{\linewidth}{!}{
        \begin{tabular}{ccccccccc}
            \toprule
            Dataset & train/test & MP & task & w/o defense & DP & GC & RLE & MLE \\
            
            \midrule
		\multirow{4}{*}{Boston Housing} & \multirow{2}{*}{train} & \multirow{2}{*}{0.7161/0.9569} & original & 0.0908/0.0182 & 0.4181/0.3102 & 0.3678/0.2579 & 0.4073/0.3048 & \textbf{0.0979/0.0183} \\
            \cmidrule(l){4-9}
            & & & attack & 0.1921/0.0688 & 0.5963/0.7169 & 0.4949/0.5235 & \textbf{0.7553/1.2007} & 0.7169/0.9707 \\
            \cmidrule(l){2-9}
            & \multirow{2}{*}{test} & \multirow{2}{*}{0.7527/1.1707} & original & 0.2108/0.1481 & 0.4332/0.3679 & 0.3957/0.3425 & 0.4293/0.3221 & \textbf{0.2376/0.1576} \\
            \cmidrule(l){4-9}
            & & & attack & 0.3033/0.1807 & 0.6194/0.7863 & 0.5321/0.5943 & \textbf{0.7867/1.2547} & 0.7534/1.1923 \\
            \midrule

		\multirow{4}{*}{California Housing} & \multirow{2}{*}{train} & \multirow{2}{*}{0.7925/1.0034} & original & 0.2534/0.1315 & 0.4854/0.4949 & 0.4536/0.4247 & 0.4239/0.3468 & \textbf{0.2575/0.1387} \\
            \cmidrule(l){4-9}
            & & & attack & 0.3379/0.2103 & 0.6457/0.8532 & 0.6196/0.7429 & \textbf{0.8297/1.0842} & 0.8144/1.0482 \\
            \cmidrule(l){2-9}
            & \multirow{2}{*}{test} & \multirow{2}{*}{0.7801/0.9851} & original & 0.2956/0.2019 & 0.5142/0.5567  & 0.4683/0.5104 & 0.4455/0.3664 & \textbf{0.2965/0.2071} \\
            \cmidrule(l){4-9}
            & & & attack & 0.3654/0.2522 & 0.6676/0.9174 & 0.6325/0.8353 & \textbf{0.8267/1.0341} & 0.8063/1.0234 \\
            \midrule
            
		\multirow{4}{*}{Power Plant} & \multirow{2}{*}{train} & \multirow{2}{*}{0.8664/0.9927} & original & 0.1563/0.0449 & 0.3942/0.3189 & 0.3478/0.2663 & 0.3385/0.1779 & \textbf{0.1616/0.0501} \\
            \cmidrule(l){4-9}
            & & & attack & 0.2895/0.1338 & 0.7293/0.9918 & 0.6646/0.9142 & \textbf{0.9025/1.1066} & 0.8715/1.0342 \\
            \cmidrule(l){2-9}
            & \multirow{2}{*}{test} & \multirow{2}{*}{0.8813/1.0293} & original & 0.1718/0.0511 & 0.4170/0.3576 & 0.3751/0.3035 & 0.3484/0.1869 & \textbf{0.1798/0.0529} \\
            \cmidrule(l){4-9}
            & & & attack & 0.2997/0.1507 & 0.7739/1.1907 & 0.7163/1.0452 & \textbf{0.9177/1.1469} & 0.8961/1.0894 \\ 
		\bottomrule
	\end{tabular}
    }
    \caption{Comparison of different defense methods. The experimental results are presented in the form of ``MAE/MSE''. ``MP'' stands for mean value prediction, ``DP'' stands for differential privacy defense method, and ``GC'' stands for gradient compression defense method. For the original task, the best performance among all defense methods is marked in bold; for the attack task, the worst attack performance among all defense methods is marked in bold.} 
    \label{table:results}
\end{table*}

\subsection{Experimental Setup}
\subsubsection{Datasets and Model Architectures}
In the experiments conducted in this paper, we utilize various datasets on regression problem, including Boston housing price, California housing price, and power plant electricity generation \cite{tufekci2014prediction}.
The two datasets, Boston housing price and California housing price, aim to predict housing price by analyzing a set of features.
The power plant dataset aims to predict the electricity generation capacity of fully loaded power plants.
We standardize each dataset and divide each dataset into training and testing sets with a ratio of 4:1.
The size and number of features for each dataset, as well as their corresponding bottom model architectures and top model architectures, are presented in \Cref{table:datasets}, where FC-$n$ refers to an $n$-layer fully connected neural network.

\subsubsection{Hyperparameters}
During the training process of the original task, we employ mean squared error (MSE) as the loss function, the Adam algorithm as the optimizer, and set the learning rate to 0.01.
To achieve good performance for the original task, we set the training epochs to 100, and set the batch size for California housing price and power plant datasets to 128, the batch size for Boston housing price dataset to 16.
During the training process of the attack task, we employ the same optimizer and learning rate as those used in the original task.
The loss function is shown in \Cref{equation:loss}, where $\alpha$ is set to $0.05$.
The batch size for the attack task is set to the same value as that of the original task, and the number of epochs for the attack training is set to 50.
Besides, the amount of leaked data is set to $1\%$ of the training dataset.

\subsubsection{Evaluation Method}
In the experiments conducted in this paper, we employ mean absolute error (MAE) and mean squared error (MSE) as the performance evaluation metrics, with MAE as the primary metric.
For both the original task and the attack task, we conduct 10 repeated experiments and select the best results as the final experimental results.

\subsection{Experimental Results}
\subsubsection{Defense Results}
\Cref{table:results} presents the defense results of different defense methods against label inference attacks in Split Learning under regression setting on various datasets.
``MP'' refers to mean value prediction, which utilizes the mean value of the labels in a dataset as the prediction.
When the performance of a model's prediction is worse than that of mean value prediction, the model is generally considered to be ineffective.
``DP'' refers to differential privacy defense method, and experiments in this paper involve adding Laplace noise to the labels.
Adding noise to the gradients or utilizing other types of noise can yield similar outcomes.
In \Cref{table:results}, the scale for Laplace noise is set to $1$.
``GC'' refers to gradient compression defense method, and gradient compression in the experiments is implemented through gradient sparsification.
In \Cref{table:results}, the compression ratio for gradient sparsification is set to $50\%$.
Besides, for both RLE and MLE, the labels are extended to $D_E$ dimensions. 
According to the experimental results presented in \Cref{table:results}, the attack method utilized in this paper can effectively conduct label inference attacks in Split Learning under regression setting in the absence of defense methods.
Basic defense methods such as adding noise and gradient compression can achieve a certain degree of defense effectiveness.
However, the defense performance of basic defense methods is evidently inferior to that of RLE.
Additionally, the preservation of original task's performance when basic defense methods are applied is also generally inferior to that when RLE is applied.
The analysis of basic defense methods using other parameters will be presented in the next section.
RLE achieves the best defense performance against label inference attacks among these defense methods;
however, it still leads to significant performance deterioration in the original task.
MLE can effectively defend against label inference attacks while maximizing the preservation of the original task's performance.
Actually, the difference in the original task's performance between applying MLE and no defense method is negligible and can be essentially disregarded.
According to \Cref{table:results}, despite the defense performance of MLE being inferior to that of RLE, the attack model's performance when MLE is applied is already worse than that when mean value prediction is applied.
Therefore, we consider MLE's defense performance to be sufficiently effective.

\subsubsection{Comparison with Baselines}
\begin{figure}[t]
    \centering
    \begin{subfigure}{1\linewidth}
        \centering
        \includegraphics[width=0.6\linewidth]{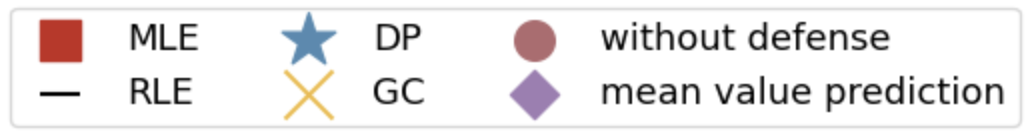}
        \label{fig:legend}
    \end{subfigure}
    \begin{subfigure}{0.495\linewidth}
        \includegraphics[width=1\linewidth]{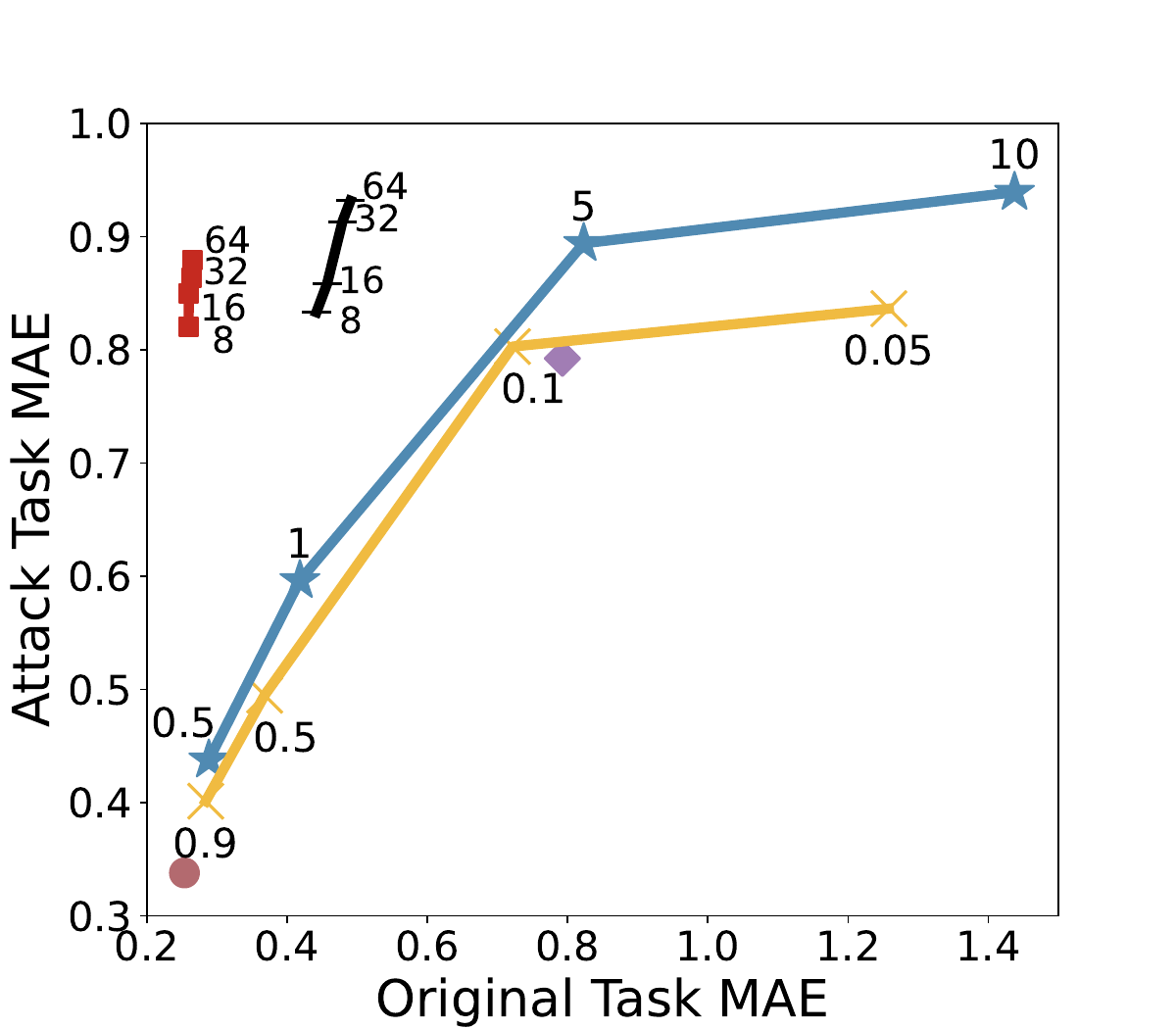}
        \subcaption{California Housing (train).}
        \label{fig:Ctrain}
    \end{subfigure}
    \begin{subfigure}{0.495\linewidth}
        \includegraphics[width=1\linewidth]{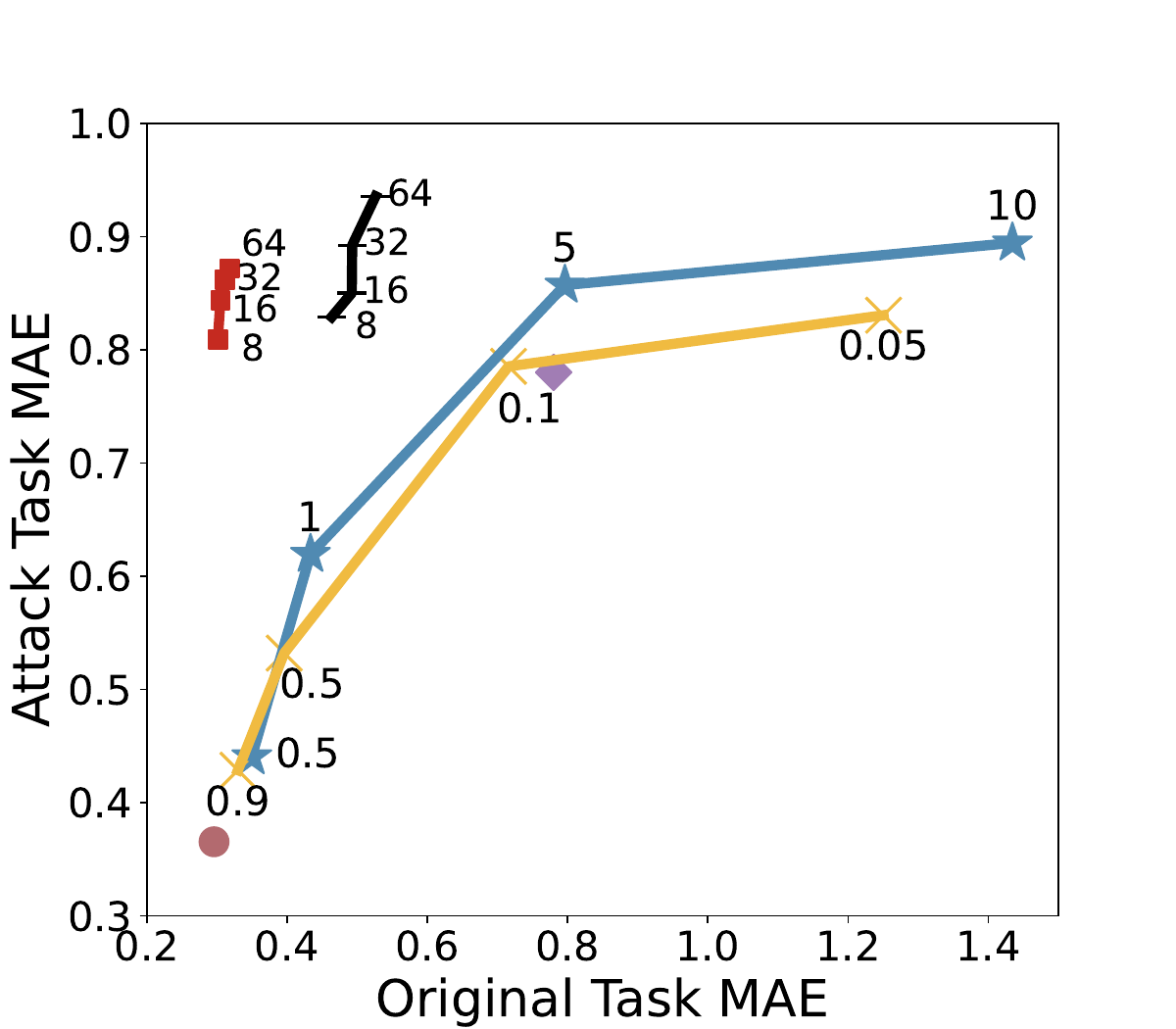}
        \subcaption{California Housing (test).}
        \label{fig:Ctest}
    \end{subfigure}
    \begin{subfigure}{0.495\linewidth}
        \includegraphics[width=1\linewidth]{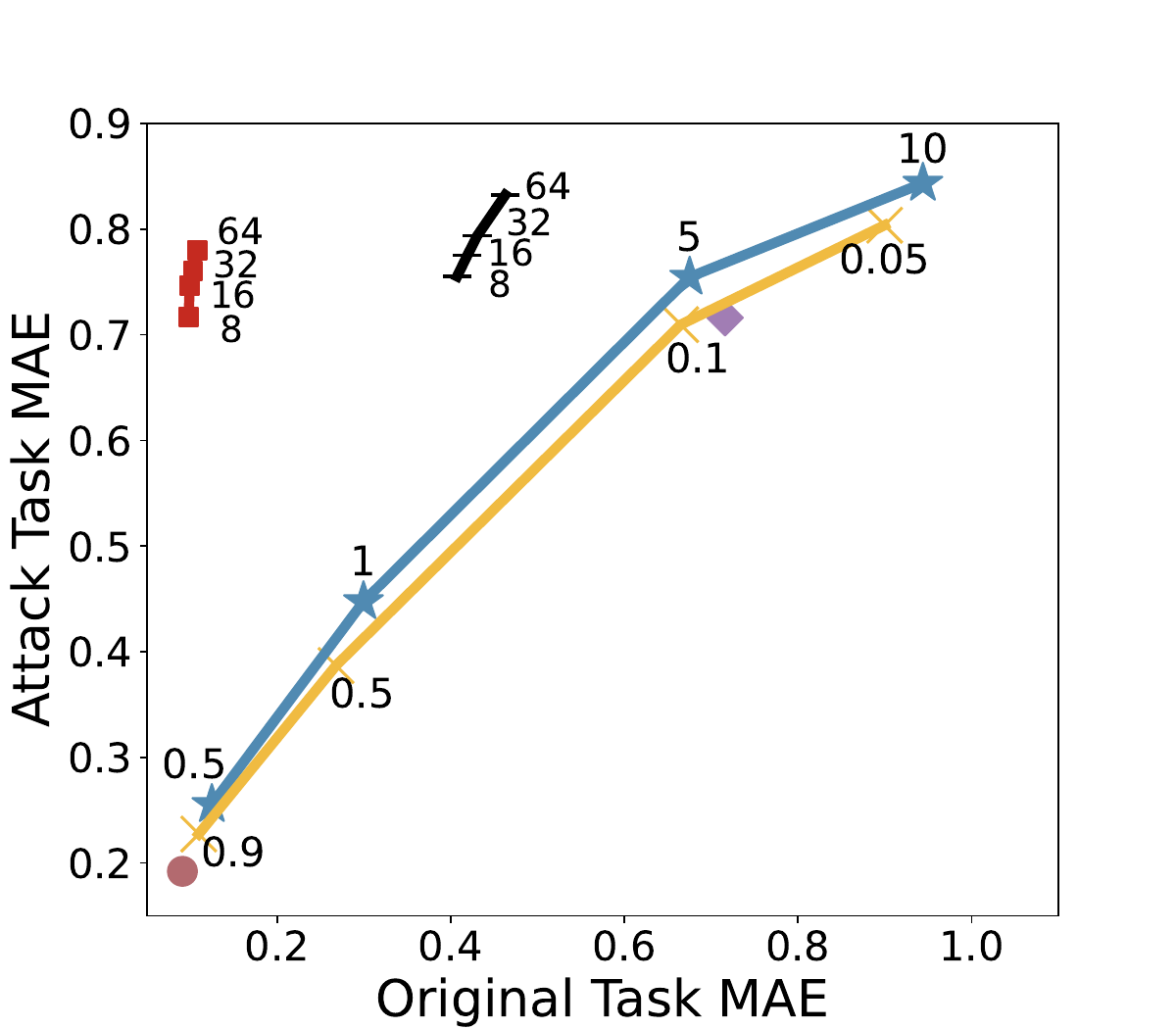}
        \subcaption{Boston Housing (train).}
        \label{fig:Btrain}
    \end{subfigure}
    \begin{subfigure}{0.495\linewidth}
        \includegraphics[width=1\linewidth]{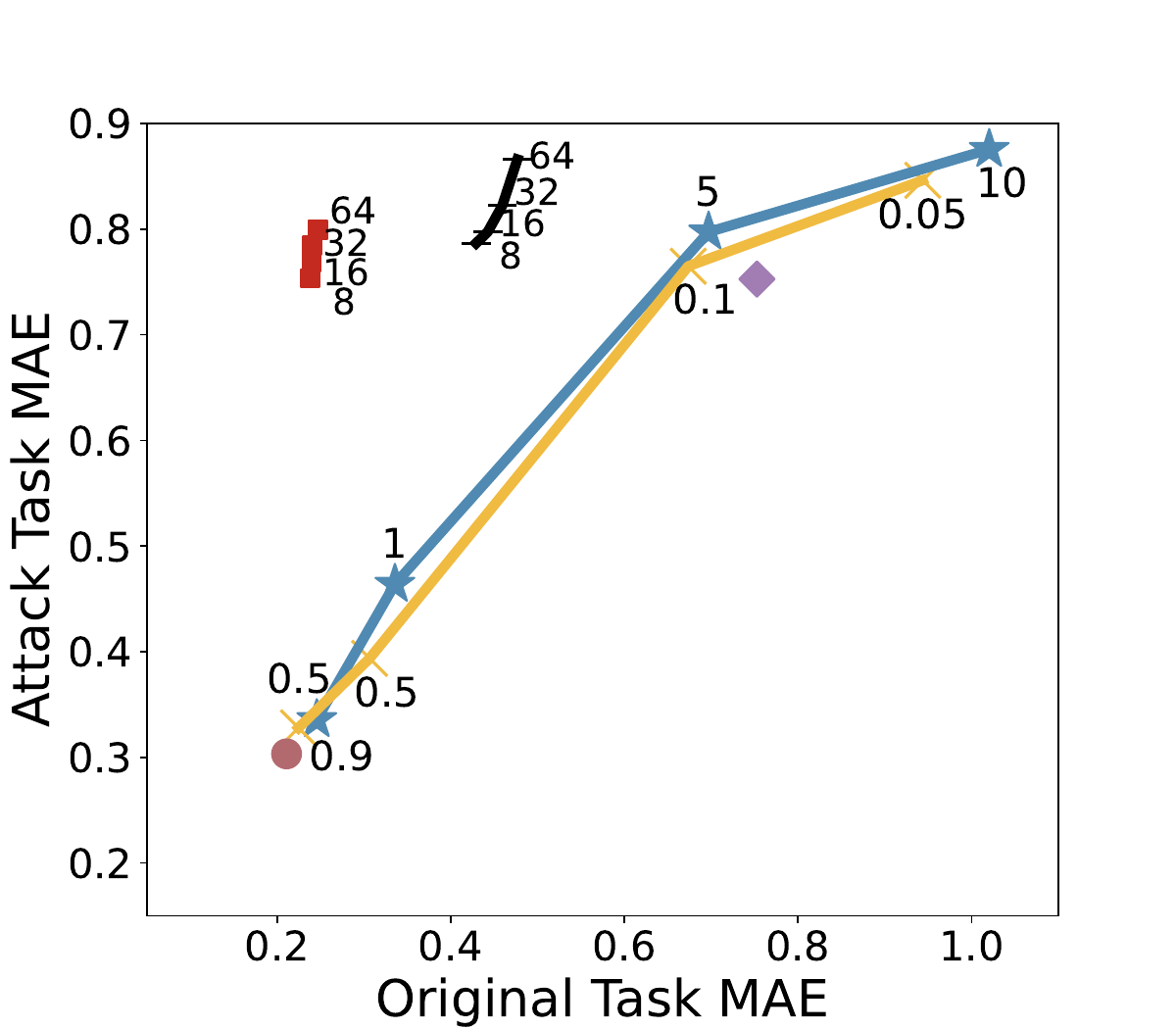}
        \subcaption{Boston Housing (test).}
        \label{fig:Btest}
    \end{subfigure}  
    \begin{subfigure}{0.495\linewidth}
        \includegraphics[width=1\linewidth]{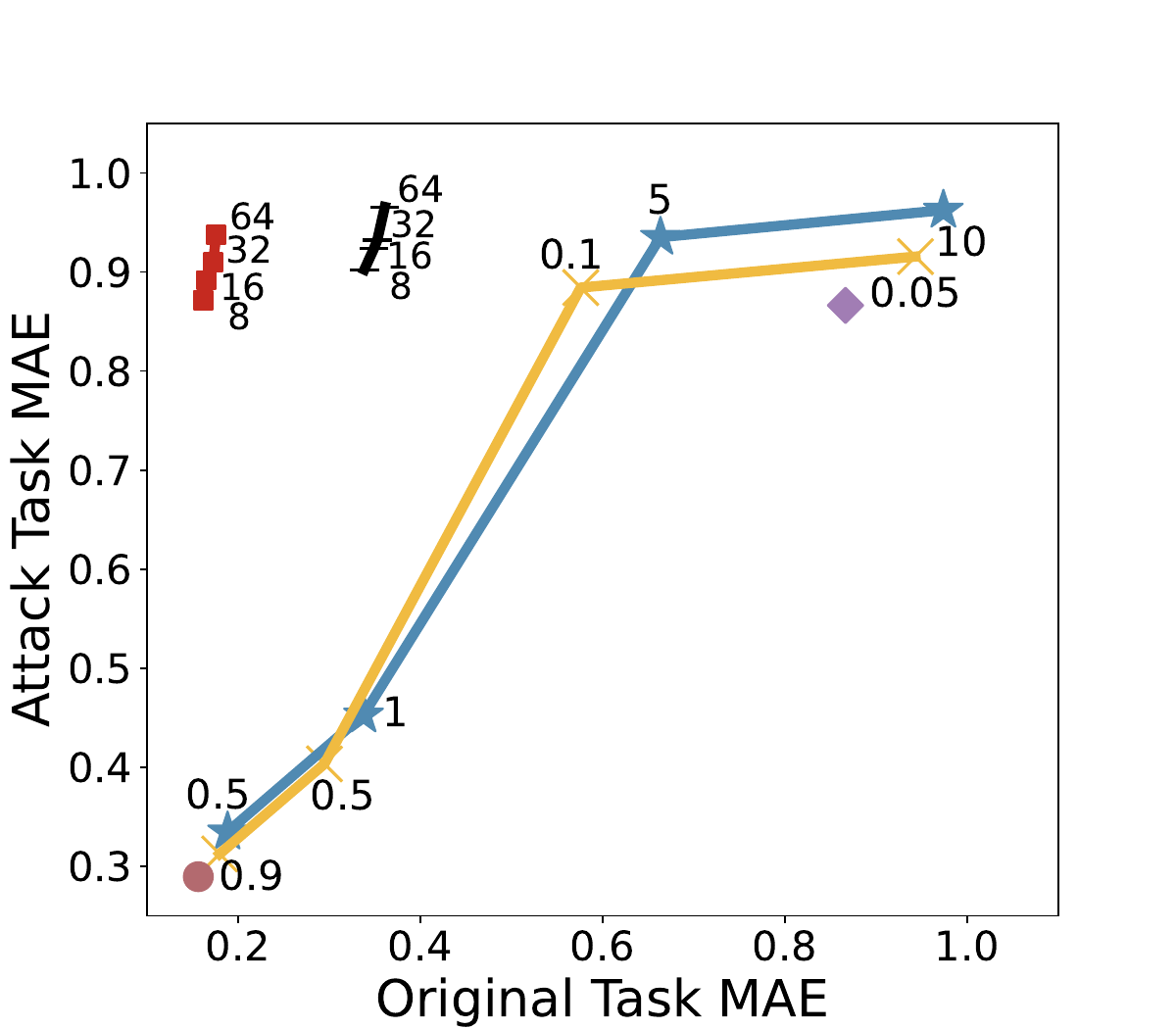}
        \subcaption{Power Plant (train).}
        \label{fig:Ptrain}
    \end{subfigure}    
    \begin{subfigure}{0.495\linewidth}
        \includegraphics[width=1\linewidth]{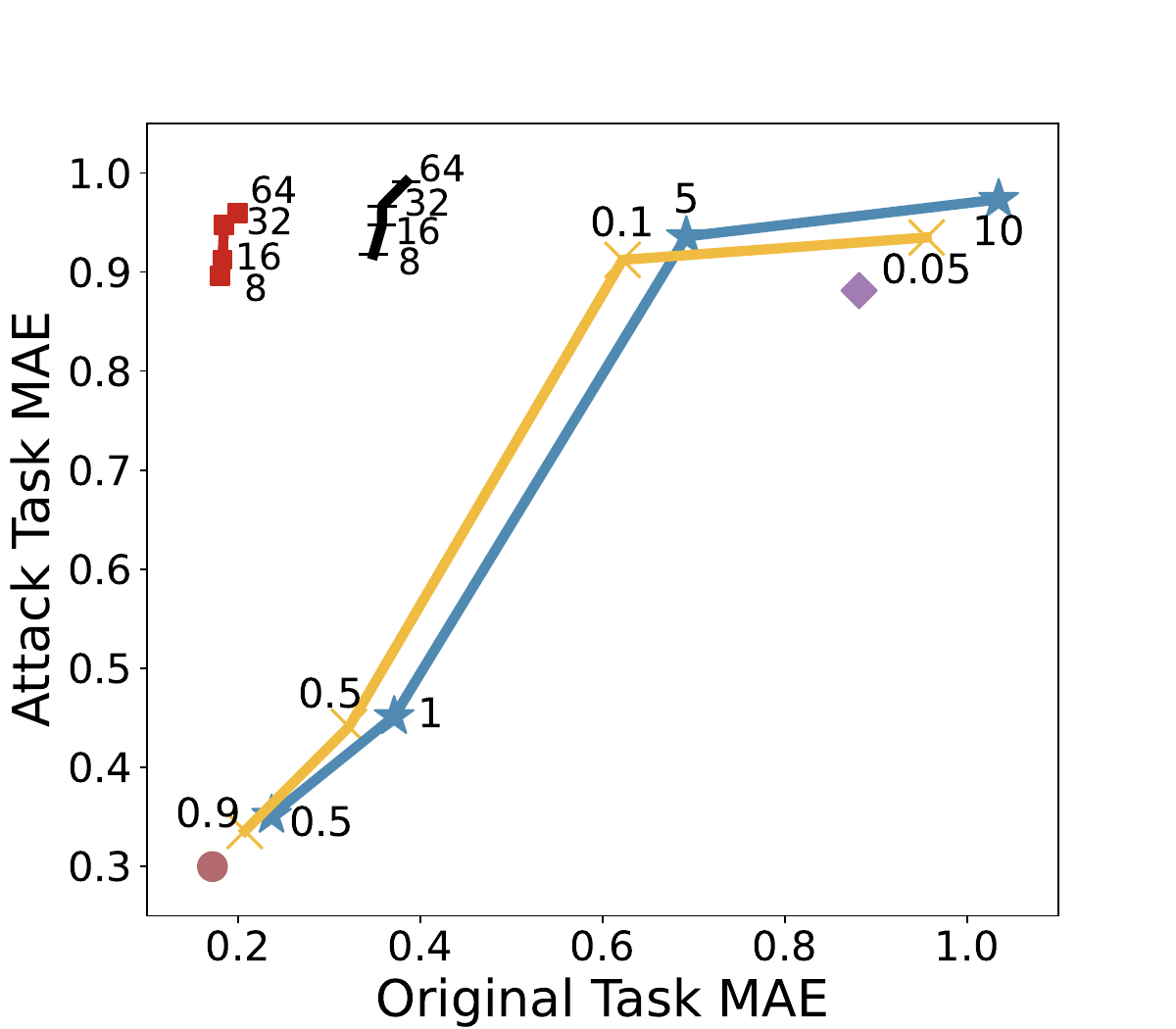}
        \subcaption{Power Plant (test).}
        \label{fig:Ptest}
    \end{subfigure} 
    \caption{MAE of the original task vs. MAE of the attack task under various defense strategies on regression tasks. The numbers in figures represent the parameters of various defense methods.}
    \label{fig:oa}
\end{figure}

We further compare the defense effect of RLE, MLE, and basic defense methods such as adding noise and gradient compression with different parameters.
The results are shown in \Cref{fig:oa}.
The horizontal axis in the figure represents the MAE of the original task while the vertical axis represents the MAE of the attack task.
The numbers in the figure represent the parameters of various defense methods. 
For adding noise, the numbers represent the noise scale; for gradient compression, the numbers represent the compression rate; for MLE and RLE, the numbers represent the dimension of extended labels.
According to \Cref{fig:oa}, it can be concluded that the defense methods of adding noise or gradient compression cannot balance effective defense and preservation of the original task's performance, regardless of how the noise scale or the compression rate is adjusted.
When the noise scale is small or the compression rate is high, the original task's performance is high, but the attack task's performance is also high, indicating that the defense has little effect.
When the noise scale is large or the compression rate is low, the attack task's performance is reduced, but the original task's performance also becomes very low, which renders the defense meaningless.
Compared with basic defense methods, RLE can effectively defend against label inference attacks while preserving the original task's performance to a certain extent.
MLE can ensure that the difference between the original task's performance when applying MLE and the original task's performance without defense to be negligible, while achieving effective defense.

\subsubsection{Impact of Different Extension Dimensions}
\begin{figure}[t]
    \centering
    \begin{subfigure}{0.49\linewidth}
        \includegraphics[width=1\linewidth]{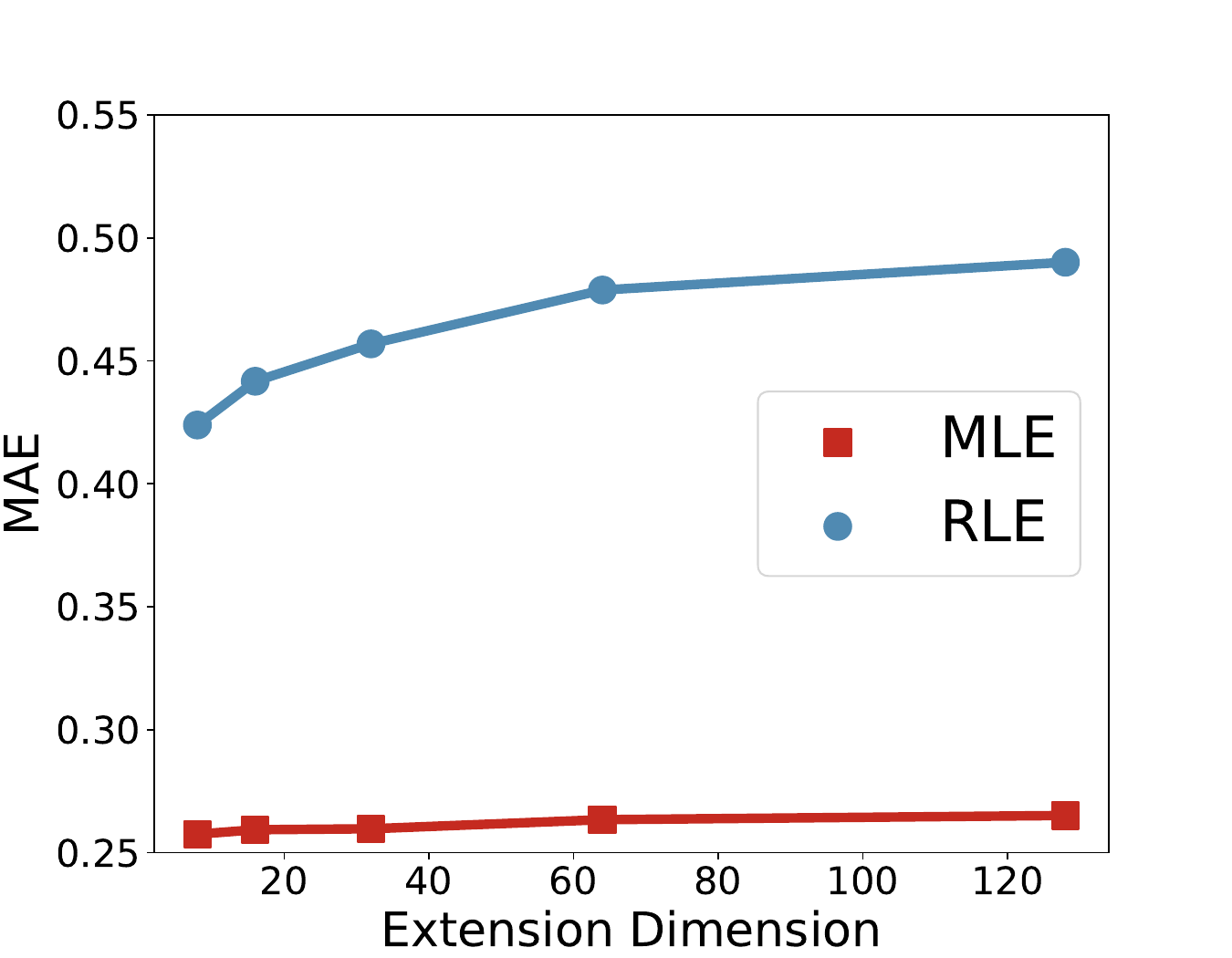}
        \subcaption{Original task (train).}
        \label{fig:Otrain}
    \end{subfigure}  
    \begin{subfigure}{0.49\linewidth}
        \includegraphics[width=1\linewidth]{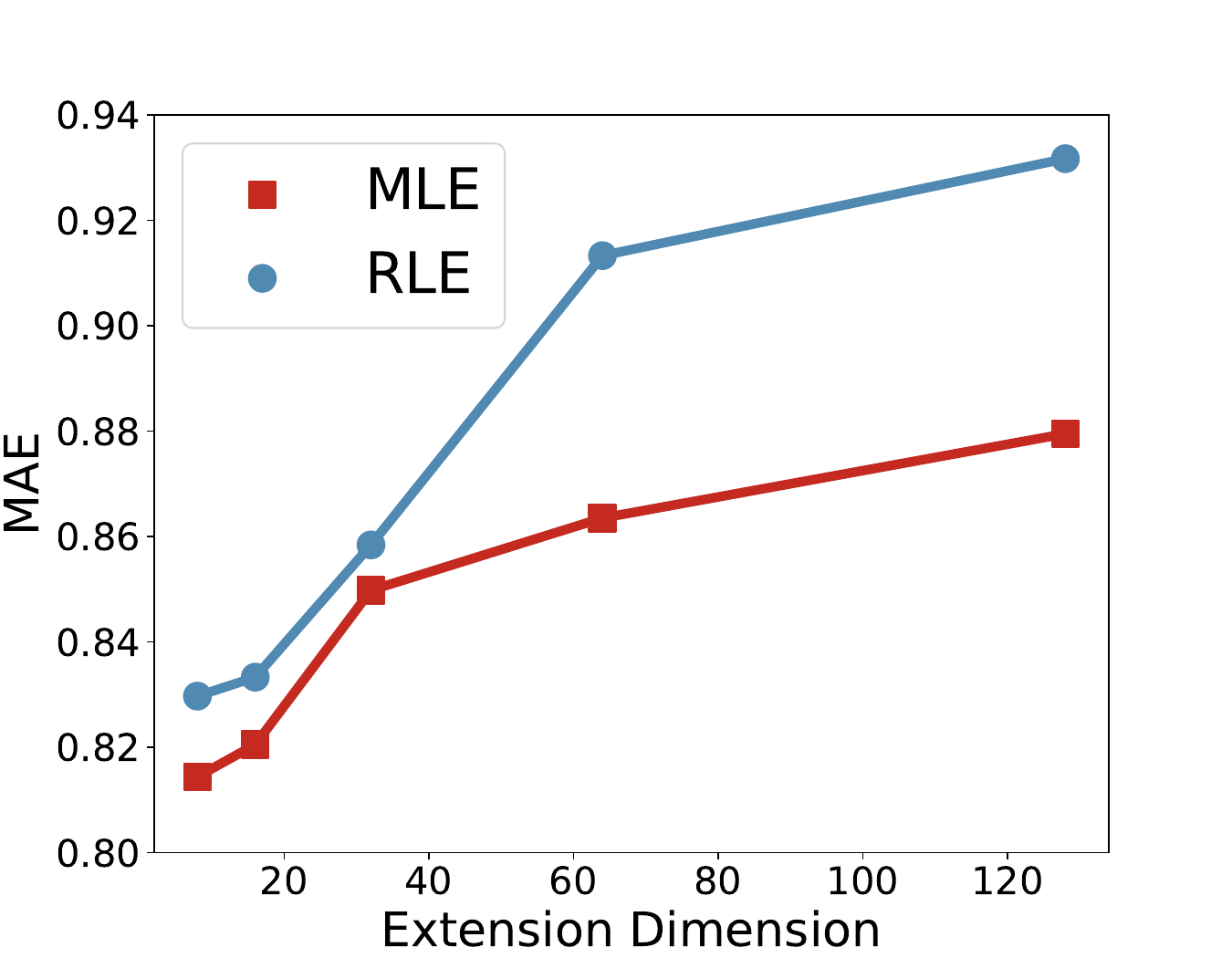}
        \subcaption{Attack task (train).}
        \label{fig:Atrain}
    \end{subfigure}   
    \begin{subfigure}{0.49\linewidth}
        \includegraphics[width=1\linewidth]{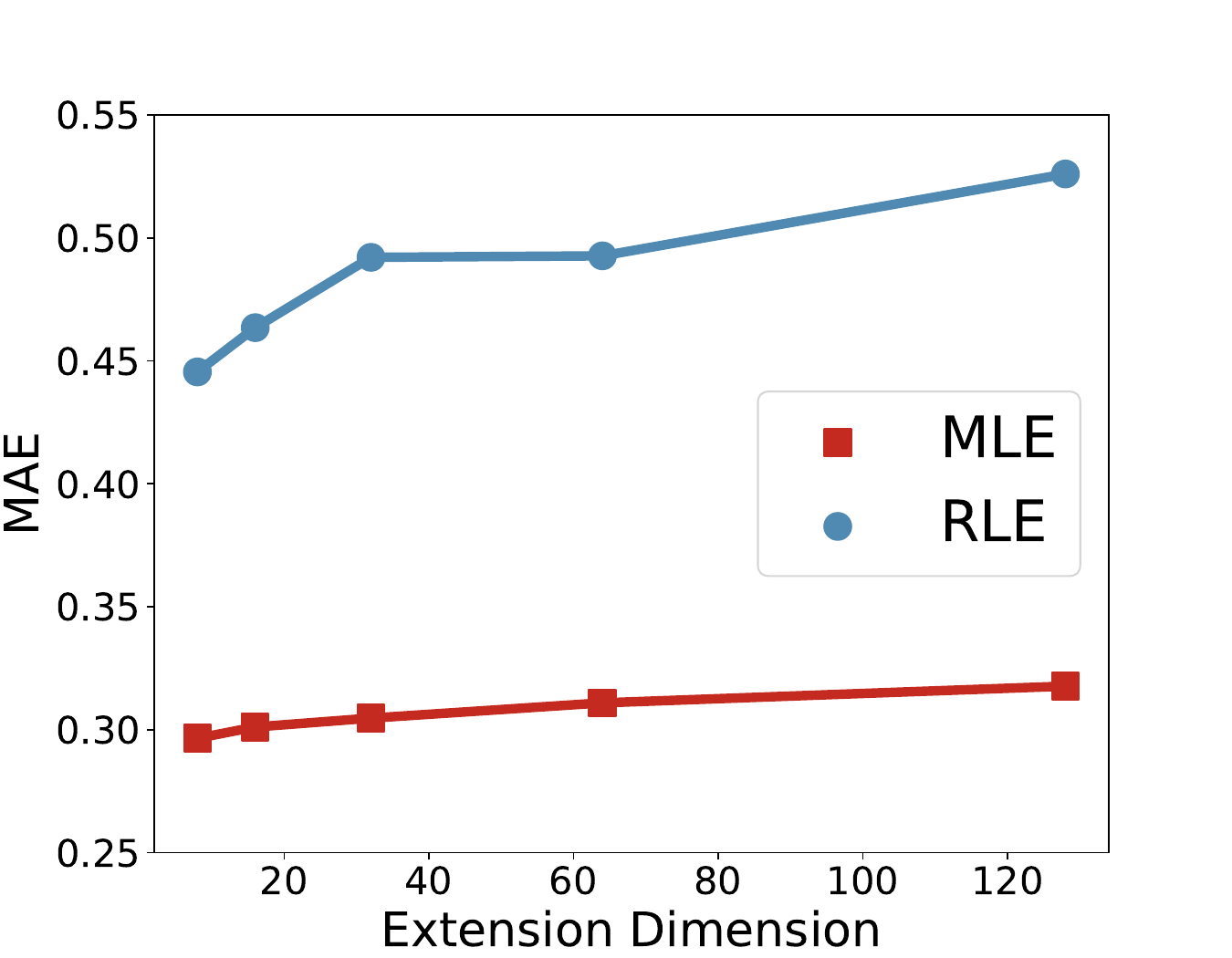}
        \subcaption{Original task (test).}
        \label{fig:Otest}
    \end{subfigure}    
    \begin{subfigure}{0.49\linewidth}
        \includegraphics[width=1\linewidth]{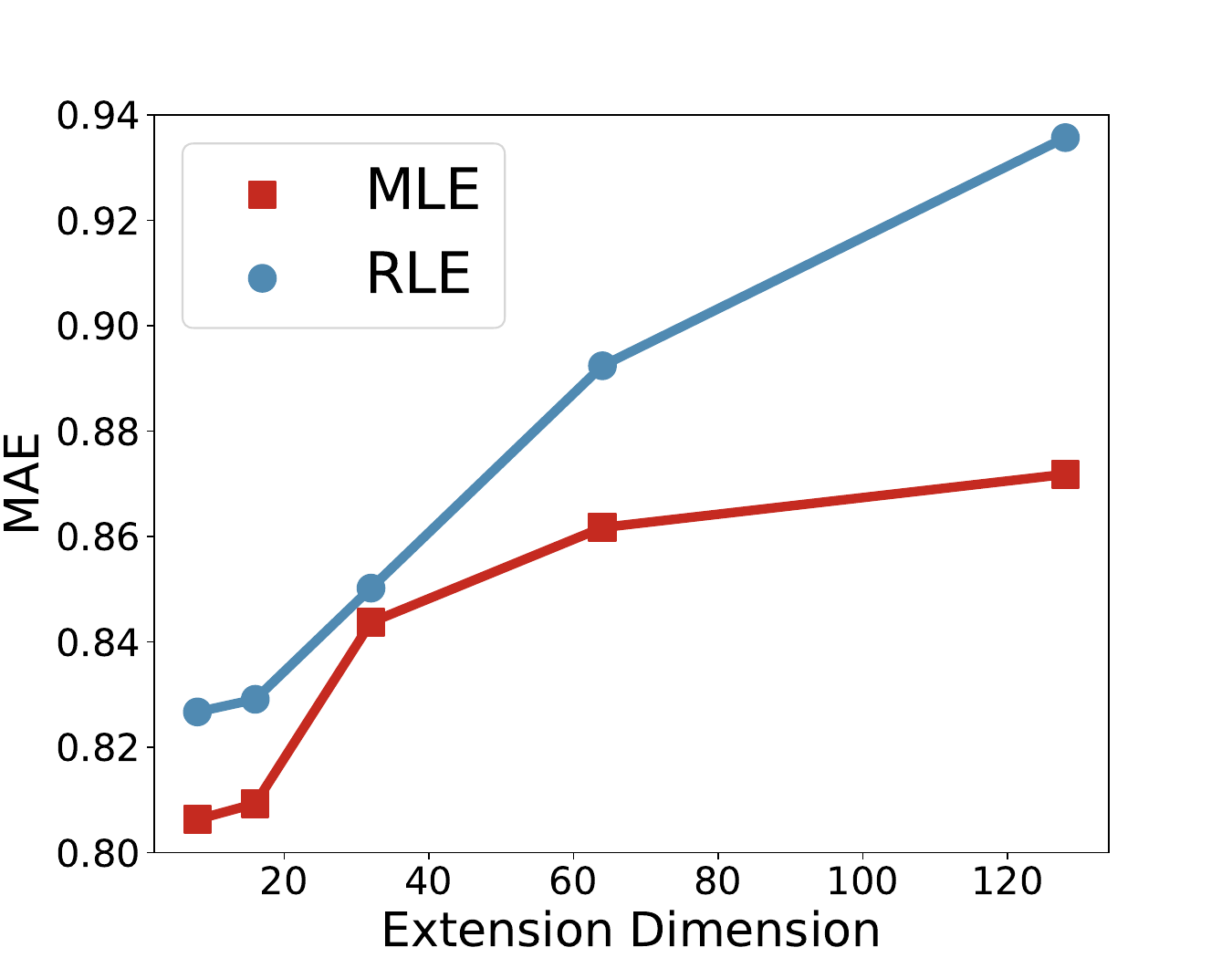}
        \subcaption{Attack task (test).}
        \label{fig:Atest}
    \end{subfigure}    
    \caption{MAE of the original task and the attack task under different extension dimensions (California Housing). }
    \label{fig:extension_dimensions}
\end{figure}

We have studied the impact of different extension dimensions on the original task and the attack task for RLE and MLE, and the results are shown in \Cref{fig:extension_dimensions}.
For the original task, as the extension dimension increases, the original task's performance decreases.
Besides, the original task's performance is higher when applying MLE, and as the extension dimension increases, the original task's performance decreases more slowly when applying MLE.
For the attack task, as the extension dimension increases, the attack task's performance decreases.
The attack task's performance is worse when applying RLE, but even when extending labels to only $D_E$ dimensions, the attack task's performance when applying MLE is already worse than the performance of mean value prediction, indicating that the defense is effective enough.
Conclusively, MLE is the most suitable defense method.
%


\section{Conclusion}
In this paper, we investigate defense methods against label inference attacks in Split Learning under regression setting.
We find that the label extension method can obfuscate the label information contained in the gradients, thereby preventing the attacker from using gradients to train an attack model that can infer the original labels.
Additionally, we preserve the original labels in the extended labels and enable the dimension where the original labels are located to dominate the training process, with the aim of preserving the original task's performance while effectively defending against label inference attacks.
Experiments on multiple datasets demonstrate that our proposed defense method can significantly reduce the attack model's performance with little to no impact on the original task's performance.
We believe that our proposed defense method can motivate more research on attack and defense in Split Learning under regression setting.

\bibliography{aaai24}

\begin{thebibliography}{29}
\providecommand{\natexlab}[1]{#1}

\bibitem[{Abuadbba et~al.(2020)Abuadbba, Kim, Kim, Thapa, Camtepe, Gao, Kim,
  and Nepal}]{abuadbba2020can}
Abuadbba, S.; Kim, K.; Kim, M.; Thapa, C.; Camtepe, S.~A.; Gao, Y.; Kim, H.;
  and Nepal, S. 2020.
\newblock Can we use split learning on 1d cnn models for privacy preserving
  training?
\newblock In \emph{Proceedings of the 15th ACM Asia Conference on Computer and
  Communications Security}, 305--318.

\bibitem[{Aggarwal et~al.(2021)Aggarwal, Kasiviswanathan, Xu, Feyisetan, and
  Teissier}]{aggarwal2021label}
Aggarwal, A.; Kasiviswanathan, S.; Xu, Z.; Feyisetan, O.; and Teissier, N.
  2021.
\newblock Label inference attacks from log-loss scores.
\newblock In \emph{International Conference on Machine Learning}, 120--129.
  PMLR.

\bibitem[{Castiglia et~al.(2022)Castiglia, Das, Wang, and
  Patterson}]{castiglia2022compressed}
Castiglia, T.~J.; Das, A.; Wang, S.; and Patterson, S. 2022.
\newblock Compressed-vfl: Communication-efficient learning with vertically
  partitioned data.
\newblock In \emph{International Conference on Machine Learning}, 2738--2766.
  PMLR.

\bibitem[{Chen et~al.(2022)Chen, Wu, Su, Lyu, Zheng, and
  Wang}]{chen2022differential}
Chen, C.; Wu, H.; Su, J.; Lyu, L.; Zheng, X.; and Wang, L. 2022.
\newblock Differential private knowledge transfer for privacy-preserving
  cross-domain recommendation.
\newblock In \emph{Proceedings of the ACM Web Conference 2022}, 1455--1465.

\bibitem[{Chen, Li, and Chakrabarti(2021)}]{chen2021communication}
Chen, X.; Li, J.; and Chakrabarti, C. 2021.
\newblock Communication and computation reduction for split learning using
  asynchronous training.
\newblock In \emph{2021 IEEE Workshop on Signal Processing Systems (SiPS)},
  76--81. IEEE.

\bibitem[{Dwork et~al.(2006)Dwork, McSherry, Nissim, and
  Smith}]{dwork2006calibrating}
Dwork, C.; McSherry, F.; Nissim, K.; and Smith, A. 2006.
\newblock Calibrating noise to sensitivity in private data analysis.
\newblock In \emph{Theory of Cryptography: Third Theory of Cryptography
  Conference, TCC 2006, New York, NY, USA, March 4-7, 2006. Proceedings 3},
  265--284. Springer.

\bibitem[{Dwork, Roth et~al.(2014)}]{dwork2014algorithmic}
Dwork, C.; Roth, A.; et~al. 2014.
\newblock The algorithmic foundations of differential privacy.
\newblock \emph{Foundations and Trends{\textregistered} in Theoretical Computer
  Science}, 9(3--4): 211--407.

\bibitem[{Fu et~al.(2022{\natexlab{a}})Fu, Zhang, Ji, Chen, Wu, Guo, Zhou, Liu,
  and Wang}]{fu2022label}
Fu, C.; Zhang, X.; Ji, S.; Chen, J.; Wu, J.; Guo, S.; Zhou, J.; Liu, A.~X.; and
  Wang, T. 2022{\natexlab{a}}.
\newblock Label inference attacks against vertical federated learning.
\newblock In \emph{31st USENIX Security Symposium (USENIX Security 22)},
  1397--1414.

\bibitem[{Fu et~al.(2022{\natexlab{b}})Fu, Xue, Cheng, Tao, and
  Cui}]{fu2022blindfl}
Fu, F.; Xue, H.; Cheng, Y.; Tao, Y.; and Cui, B. 2022{\natexlab{b}}.
\newblock Blindfl: Vertical federated machine learning without peeking into
  your data.
\newblock In \emph{Proceedings of the 2022 International Conference on
  Management of Data}, 1316--1330.

\bibitem[{Ghazi et~al.(2021)Ghazi, Golowich, Kumar, Manurangsi, and
  Zhang}]{ghazi2021deep}
Ghazi, B.; Golowich, N.; Kumar, R.; Manurangsi, P.; and Zhang, C. 2021.
\newblock Deep learning with label differential privacy.
\newblock \emph{Advances in neural information processing systems}, 34:
  27131--27145.

\bibitem[{Hu et~al.(2020)Hu, Guo, Li, Pei, and Gong}]{hu2020personalized}
Hu, R.; Guo, Y.; Li, H.; Pei, Q.; and Gong, Y. 2020.
\newblock Personalized federated learning with differential privacy.
\newblock \emph{IEEE Internet of Things Journal}, 7(10): 9530--9539.

\bibitem[{Huang et~al.(2022)Huang, Lu, Hong, and Ding}]{huang2022cheetah}
Huang, Z.; Lu, W.-j.; Hong, C.; and Ding, J. 2022.
\newblock Cheetah: Lean and fast secure $\{$two-party$\}$ deep neural network
  inference.
\newblock In \emph{31st USENIX Security Symposium (USENIX Security 22)},
  809--826.

\bibitem[{Li et~al.(2021)Li, Sun, Yang, Gao, Zhang, Xie, Smith, and
  Wang}]{li2021label}
Li, O.; Sun, J.; Yang, X.; Gao, W.; Zhang, H.; Xie, J.; Smith, V.; and Wang, C.
  2021.
\newblock Label Leakage and Protection in Two-party Split Learning.
\newblock In \emph{International Conference on Learning Representations}.

\bibitem[{Liu and Lyu(2022)}]{liu2022clustering}
Liu, J.; and Lyu, X. 2022.
\newblock Clustering label inference attack against practical split learning.
\newblock \emph{arXiv preprint arXiv:2203.05222}.

\bibitem[{Luo et~al.(2021)Luo, Wu, Xiao, and Ooi}]{luo2021feature}
Luo, X.; Wu, Y.; Xiao, X.; and Ooi, B.~C. 2021.
\newblock Feature inference attack on model predictions in vertical federated
  learning.
\newblock In \emph{2021 IEEE 37th International Conference on Data Engineering
  (ICDE)}, 181--192. IEEE.

\bibitem[{Mohassel and Zhang(2017)}]{mohassel2017secureml}
Mohassel, P.; and Zhang, Y. 2017.
\newblock Secureml: A system for scalable privacy-preserving machine learning.
\newblock In \emph{2017 IEEE symposium on security and privacy (SP)}, 19--38.
  IEEE.

\bibitem[{Pasquini, Ateniese, and Bernaschi(2021)}]{pasquini2021unleashing}
Pasquini, D.; Ateniese, G.; and Bernaschi, M. 2021.
\newblock Unleashing the tiger: Inference attacks on split learning.
\newblock In \emph{Proceedings of the 2021 ACM SIGSAC Conference on Computer
  and Communications Security}, 2113--2129.

\bibitem[{Rathee et~al.(2020)Rathee, Rathee, Kumar, Chandran, Gupta, Rastogi,
  and Sharma}]{rathee2020cryptflow2}
Rathee, D.; Rathee, M.; Kumar, N.; Chandran, N.; Gupta, D.; Rastogi, A.; and
  Sharma, R. 2020.
\newblock Cryptflow2: Practical 2-party secure inference.
\newblock In \emph{Proceedings of the 2020 ACM SIGSAC Conference on Computer
  and Communications Security}, 325--342.

\bibitem[{Sun et~al.(2022)Sun, Yang, Yao, and Wang}]{sun2022label}
Sun, J.; Yang, X.; Yao, Y.; and Wang, C. 2022.
\newblock Label leakage and protection from forward embedding in vertical
  federated learning.
\newblock \emph{arXiv preprint arXiv:2203.01451}.

\bibitem[{T{\"u}fekci(2014)}]{tufekci2014prediction}
T{\"u}fekci, P. 2014.
\newblock Prediction of full load electrical power output of a base load
  operated combined cycle power plant using machine learning methods.
\newblock \emph{International Journal of Electrical Power \& Energy Systems},
  60: 126--140.

\bibitem[{Vepakomma et~al.(2018)Vepakomma, Gupta, Swedish, and
  Raskar}]{vepakomma2018split}
Vepakomma, P.; Gupta, O.; Swedish, T.; and Raskar, R. 2018.
\newblock Split learning for health: Distributed deep learning without sharing
  raw patient data.
\newblock \emph{arXiv preprint arXiv:1812.00564}.

\bibitem[{Vepakomma et~al.(2020)Vepakomma, Singh, Gupta, and
  Raskar}]{vepakomma2020nopeek}
Vepakomma, P.; Singh, A.; Gupta, O.; and Raskar, R. 2020.
\newblock NoPeek: Information leakage reduction to share activations in
  distributed deep learning.
\newblock In \emph{2020 International Conference on Data Mining Workshops
  (ICDMW)}, 933--942. IEEE.

\bibitem[{Wei et~al.(2020)Wei, Li, Ding, Ma, Yang, Farokhi, Jin, Quek, and
  Poor}]{wei2020federated}
Wei, K.; Li, J.; Ding, M.; Ma, C.; Yang, H.~H.; Farokhi, F.; Jin, S.; Quek,
  T.~Q.; and Poor, H.~V. 2020.
\newblock Federated learning with differential privacy: Algorithms and
  performance analysis.
\newblock \emph{IEEE Transactions on Information Forensics and Security}, 15:
  3454--3469.

\bibitem[{Wu et~al.(2022)Wu, Zhou, Weinberger, and Guo}]{wu2022does}
Wu, R.; Zhou, J.~P.; Weinberger, K.~Q.; and Guo, C. 2022.
\newblock Does Label Differential Privacy Prevent Label Inference Attacks?
\newblock \emph{arXiv preprint arXiv:2202.12968}.

\bibitem[{Xie et~al.(2023)Xie, Yang, Yao, Liu, Wang, and Sun}]{xie2023label}
Xie, S.; Yang, X.; Yao, Y.; Liu, T.; Wang, T.; and Sun, J. 2023.
\newblock Label Inference Attack against Split Learning under Regression
  Setting.
\newblock \emph{arXiv preprint arXiv:2301.07284}.

\bibitem[{Zheng et~al.(2023)Zheng, Chen, Lyu, and Yao}]{zheng2023reducing}
Zheng, F.; Chen, C.; Lyu, L.; and Yao, B. 2023.
\newblock Reducing Communication for Split Learning by Randomized Top-k
  Sparsification.
\newblock \emph{arXiv preprint arXiv:2305.18469}.

\bibitem[{Zheng et~al.(2022)Zheng, Chen, Yao, and Zheng}]{zheng2022making}
Zheng, F.; Chen, C.; Yao, B.; and Zheng, X. 2022.
\newblock Making split learning resilient to label leakage by potential energy
  loss.
\newblock \emph{arXiv preprint arXiv:2210.09617}.

\bibitem[{Zhu, Liu, and Han(2019)}]{zhu2019deep}
Zhu, L.; Liu, Z.; and Han, S. 2019.
\newblock Deep leakage from gradients.
\newblock \emph{Advances in neural information processing systems}, 32.

\bibitem[{Zou et~al.(2022)Zou, Liu, Kang, Liu, He, Yi, Yang, and
  Zhang}]{zou2022defending}
Zou, T.; Liu, Y.; Kang, Y.; Liu, W.; He, Y.; Yi, Z.; Yang, Q.; and Zhang, Y.-Q.
  2022.
\newblock Defending batch-level label inference and replacement attacks in
  vertical federated learning.
\newblock \emph{IEEE Transactions on Big Data}.

\end{thebibliography}

\end{document}